\documentclass{article}
\usepackage{iclr2021_conference,times}
\iclrfinalcopy
\usepackage[utf8]{inputenc}
\usepackage{hyperref}
\usepackage{natbib}

\title{Neural Volume Rendering: NeRF and Beyond}
\author{Frank Dellaert \\
Georgia Institute of Technology\\
\texttt{dellaert@cc.gatech.edu} \\
\And
Lin Yen-Chen \\
Massachusetts Institute of Technology \\
\texttt{yenchenl@mit.edu}
}

\begin{document}

\maketitle

\section{Introduction}

Besides the COVID-19 pandemic and political upheaval in the US, 2020 was also the year in which neural volume rendering exploded onto the scene, triggered by the impressive NeRF paper by \citet{Mildenhall20eccv_nerf}. 
Both of us have tried to capture this excitement, Frank on a \href{https://dellaert.github.io/NeRF/}{blog post}~\citep{Dellaert20post_NeRF} and Yen-Chen in a \href{https://github.com/yenchenlin/awesome-NeRF}{Github collection}~\citep{YenChen20github_awesome_NeRF}.
This note is an annotated bibliography of the relevant papers, and we posted the associated bibtex file \href{https://github.com/yenchenlin/awesome-NeRF}{on the repository}.

To start with some definitions, the larger field of \textbf{Neural Rendering} is defined by the excellent review paper by \citet{Tewari20eurographics_neural_rendering} as
\begin{quotation}
\noindent
``deep image or video generation approaches that enable explicit or implicit control of scene properties such as illumination, camera parameters, pose, geometry, appearance, and semantic structure.''
\end{quotation}
It is a novel, data-driven solution to the long-standing problem in computer graphics of the realistic rendering of virtual worlds.

\textbf{Neural volume rendering} refers to methods that generate images or video by tracing a ray into the scene and taking an integral of some sort over the length of the ray. Typically a neural network like a multi-layer perceptron encodes a function from the 3D coordinates on the ray to quantities like density and color, which are integrated to yield an image.

\paragraph{Outline} Below we first discuss some very relevant related works that lead up to the “NeRF explosion”, then discuss the two papers that we think started it all, followed by an annotated bibliography on follow-up work. We are going wide rather than deep, and provide links to all project sites or Arxiv entries. 

\section{The Prelude: Neural Implicit Surfaces}

The immediate precursors to neural volume rendering are the approaches that use a neural network to define an implicit surface representation. Many 3D-aware image generation approaches used voxels, meshes, point clouds, or other representations, typically based on convolutional architectures. But at CVPR 2019, no less than three papers introduced the use of neural nets as scalar function approximators to define occupancy and/or signed distance functions.

\subsection{Occupancy and Signed Distance Functions}

Below are the three papers from CVPR 2019, and one (PIFu) from ICCV 2019:
\begin{itemize}
\item 
\href{https://avg.is.tuebingen.mpg.de/publications/occupancy-networks}{Occupancy networks}~\citep{Mescheder19cvpr_occupancy_net} introduce implicit, coordinate-based learning of \textbf{occupancy}. A network consisting of 5 ResNet blocks take a feature vector and a 3D point and predict binary occupancy.

\item 
\href{https://github.com/czq142857/implicit-decoder}{IM-NET}~\citep{Chen19cvpr_IM_NET} uses a 6-layer MLP decoder that predicts binary occupancy given a feature vector and a 3D coordinate. Can be used for auto-encoding, shape generation (GAN-style), and single-view reconstruction.

\item 
\href{https://github.com/facebookresearch/DeepSDF}{DeepSDF}~\citep{Park19cvpr_deepsdf} directly regresses a \textbf{signed distance function} from a 3D coordinate and optionally a latent code. It uses an 8-layer MPL with skip-connections to layer 4, setting a trend!

\item 
\href{https://shunsukesaito.github.io/PIFu/}{PIFu}~\citep{Saito19Iccv_PIFu} shows that it is possible to learn highly detailed implicit models by re-projecting 3D points into a pixel-aligned feature representation. This idea will later be reprised, with great effect, in PixelNeRF.

\end{itemize}

\subsection{Building on Implicit Functions}
Several other approaches build on top of the implicit function idea.

\begin{itemize}
\item
\href{https://ldif.cs.princeton.edu/}{Structured Implicit Functions}~\citep{Genova19iccv_sif} show that you can combine these implicit representations, e.g., simply by summing them. 

\item
\href{https://cvxnet.github.io/}{CvxNet}~\citep{Deng20cvpr_cvxnet} combines signed distance functions by taking a pointwise max (in 3D). The paper also has several other elegant techniques to reconstruct an object from depth or RGB images. 

\item
\href{https://bsp-net.github.io/}{BSP-Net}~\citep{Chen20cvpr_bsp} is in many ways similar to CvxNet, but uses binary space partitioning at its core, yielding a method that outputs polygonal meshes natively, rather than via an expensive meshing method.

\item
\href{https://arxiv.org/abs/2003.10983}{Deep Local Shapes}~\citep{Chabra20eccv_DLS} store a DeepSDF latent code in a voxel grid to represent larger, extended scenes.

\item
\href{https://vsitzmann.github.io/srns/}{Scene Representation Networks}~\citep{Sitzmann19neurips_srn} or SRN are quite similar to DeepSDF in terms of architecture but adds a differentiable ray marching algorithm to find the closest point of intersection of a learned implicit surface, and add an MLP to regress color, enabling it to be learned from multiple posed images.

\item
\href{https://avg.is.tuebingen.mpg.de/publications/niemeyer2020cvpr}{Differentiable Volumetric Rendering}~\citep{Niemeyer20cvpr_DVR} shows that an implicit scene representation can be coupled with a differentiable renderer, making it trainable from images, similar to SRN. They use the term volumetric renderer, but really the main contribution is a clever trick to make the computation of depth to the implicit surface differentiable: no integration over a volume is used.

\item
\href{https://lioryariv.github.io/idr/}{Implicit Differentiable Renderer}~\citep{Yariv20neurips_MVNeuralSurfaceRecon} presents a similar technique, but has a more sophisticated surface light field representation, and also shows that it can refine camera pose during training.

\item
\href{https://virtualhumans.mpi-inf.mpg.de/nasa/}{Neural Articulated Shape Approximation}~\citep{Deng19eccv_NASA} or NASA composes implicit functions to represent articulated objects such as human bodies.
\end{itemize}

\section{Neural Volume Rendering}

As far as we know, two papers introduced \textbf{volume rendering} into the field, with NeRF being the simplest and ultimately the most influential.

\textit{A word about naming}: the two papers below and all Nerf-style papers since build upon the work above that encode implicit surfaces, and so the term implicit neural methods is used quite a bit. However, especially in graphics that term is more associated with level-set representations for curves and surfaces. What they do have in common with occupancy/SDF-style networks is that MLP’s are used as functions from coordinates in 3D to a scalar or multi-variate fields, and hence these methods are also sometimes called coordinate-based scene representation networks. Of that larger set, we’re concerned with volume rendering versions of those below.

\subsection{Neural Volumes}

While not entirely in a vacuum, we believe volume rendering for view synthesis was introduced in the \href{https://research.fb.com/publications/neural-volumes-learning-dynamic-renderable-volumes-from-images/}{Neural Volumes} paper by~\citet{Lombardi19siggraph_Neural_Volumes}, regressing a 3D volume of density and color, albeit still in a (warped) voxel-based representation. A latent code is decoded into a 3D volume, and a new image is then obtained by volume rendering.

One of the most interesting quotes from this paper hypothesizes about the success of neural volume rendering approaches (emphasis is ours):
\begin{quotation}
\noindent
[We] propose using a volumetric representation consisting of opacity and color at each position in 3D space, where rendering is realized through integral projection. During optimization, this semi-transparent representation of geometry disperses gradient information along the ray of integration, \textbf{effectively widening the basin of convergence, enabling the discovery of good solutions}.
\end{quotation}
We think that resonates with many people, and partially explains the success of neural volume rendering. We won’t go into any detail about the method itself, but the paper is a great read. Instead, let’s dive right into NeRF itself below…

\subsection{NeRF}
The paper that got everyone talking was the Neural Radiance Fields or \href{https://www.matthewtancik.com/nerf}{NeRF} paper by \citet{Mildenhall20eccv_nerf}. In essence, they take the DeepSDF architecture but regress not a signed distance function, but density and color. They then use an (easily differentiable) numerical integration method to approximate a true volumetric rendering step.

A NeRF model stores a volumetric scene representation as the weights of an MLP, trained on many images with known pose.
New views are rendered by integrating the density and color at regular intervals along each viewing ray.

One of the reasons NeRF is able to render with great detail is because it encodes a 3D point and associated view direction on a ray using periodic activation functions, i.e., Fourier Features. This innovation was later generalized to multi-layer networks with periodic activations, aka SIREN (SInusoidal REpresentation Networks). Both were published later at NeurIPS 2020.

While the NeRF paper was ostensibly published at ECCV 2020, at the end of August, it first appeared on Arxiv in the middle of March, sparking an explosion of interest, not only because of the quality of the synthesized views, but perhaps even more so at the incredible detail in the visualized depth maps.

Arguably, the impact of the NeRF paper lies in its brutal simplicity: just an MLP taking in a 5D coordinate and outputting density and color. There are some bells and whistles, notably positional encoding and a stratified sampling scheme, but many researchers were taken aback (we think) that such a simple architecture could yield such impressive results. That being said, vanilla NeRF left many opportunities to improve upon:
\begin{itemize}
\item It is slow, both for training and rendering.
\item It can only represent static scenes.
\item It “bakes in” lighting.
\item A trained NeRF representation does not generalize to other scenes/objects.
\end{itemize}
In this Arxiv-fueled computer vision world, these opportunities were almost immediately capitalized on, with almost 25 papers appearing on Arxiv in the span of six months. Below we list all of them we could find.

\section{Performance}

Several projects/papers aim at improving the rather slow training and rendering time of the original NeRF paper.

\begin{itemize}

\item 
\href{https://github.com/google-research/google-research/tree/master/jaxnerf}{JaxNeRF}~\citep{Deng20github_JaxNeRF} uses JAX (https://github.com/google/jax) to dramatically speed up training using multiple devices, from days to hours.

\item 
\href{http://www.computationalimaging.org/publications/automatic-integration/}{AutoInt}~\citep{Lindell20arxiv_AutoInt} greatly speeds up rendering by learning the volume integral directly.

\item 
\href{https://arxiv.org/abs/2012.02189}{Learned Initializations}~\citep{Tancik20arxiv_meta} uses meta-learning to find a good weight initialization for faster training.

\item 
\href{https://ubc-vision.github.io/derf/}{DeRF}~\citep{Rebain20arxiv_derf} decomposes the scene into "soft Voronoi diagrams" to take advantage of accelerator memory architectures.

\item 
\href{https://github.com/Kai-46/nerfplusplus}{NERF++}~\citep{Zhang20arxiv_nerf++} proposes to model the background with a separate NeRF to handle unbounded scenes.

\item 
\href{https://github.com/facebookresearch/NSVF}{Neural Sparse Voxel Fields}~\citep{Liu20neurips_sparse_nerf} organize the scene into a sparse voxel octree to speed up rendering by a factor of 10.

\end{itemize}

\section{Dynamic}

At least four efforts focus on dynamic scenes, using a variety of schemes.

\begin{itemize}
\item 
\href{https://nerfies.github.io/}{Nerfies}~\citep{Park20arxiv_nerfies} and its underlying D-NeRF model deformable videos using a second MLP applying a deformation for each frame of the video.

\item 
\href{https://www.albertpumarola.com/research/D-NeRF/index.html}{D-NeRF}~\citep{Pumarola20arxiv_D_NeRF} is quite similar to the Nerfies paper and even uses the same acronym, but seems to limit deformations to translations.

\item 
\href{http://www.cs.cornell.edu/~zl548/NSFF/}{Neural Scene Flow Fields}~\citep{Li20arxiv_nsff} take a monocular video with known camera poses as input but use depth predictions as a prior, and regularize by also outputting scene flow, used in the loss.

\item 
\href{https://video-nerf.github.io/}{Space-Time Neural Irradiance Fields}~\citep{Xian20arxiv_stnif} simply use time as an additional input. Carefully selected losses are needed to successfully train this method to render free-viewpoint videos (from RGBD data!).

\item 
NeRFlow~\citep{Du20arxiv_nerflow} uses a deformation MLP to model scene flow and integrates it across time to obtain the final deformation.

\item 
\href{https://gvv.mpi-inf.mpg.de/projects/nonrigid_nerf/}{NR-NeRF}~\citep{Tretschk20arxiv_NR-NeRF} also uses a deformation MLP to model non-rigid scenes. It has no reliance on pre-computed scene
information apart from camera parameters but generates slightly less sharp outputs compared to Nerfies.

\item 
\href{https://wentaoyuan.github.io/star/}{STaR}~\citep{Yuan21arxiv_star} takes multi-view RGB videos as input and decomposes the scene into a static and a dynamic volume. However, currently it only supports one object in motion.
\end{itemize}

Besides Nerfies, two other papers focus on avatars/portraits of people.

\begin{itemize}
\item 
\href{https://portrait-nerf.github.io/}{Portrait NeRF}~\citep{Gao20arxiv_pNeRF} creates static NeRF-style avatars but does so from a single RGB headshot. To make this work, light-stage training data is required.

\item 
\href{https://gafniguy.github.io/4D-Facial-Avatars/}{DNRF}~\citep{Gafni20arxiv_DNRF} focuses on 4D avatars and hence impose a strong inductive bias by including a deformable face model into the pipeline.
\end{itemize}

\section{Relighting}

Another dimension in which NeRF-style methods have been augmented is in how to deal with lighting, typically through latent codes that can be used to re-light a scene.

\begin{itemize}
\item 
\href{https://people.eecs.berkeley.edu/~pratul/nerv/}{NeRV}~\citep{Srinivasan20arxiv_NeRV} uses a second "visibility" MLP to support arbitrary environment lighting and "one-bounce" indirect illumination. 

\item 
\href{https://markboss.me/publication/2021-nerd/}{NeRD}~\citep{Boss20arxiv_NeRD} or “Neural Reflectance Decomposition” is another effort in which a local reflectance model is used, and additionally, a low-res spherical harmonics illumination is removed for a given scene. 

\item 
\href{http://cseweb.ucsd.edu/~bisai/}{Neural Reflectance Fields}~\citep{Bi20arxiv_neural_reflection_fields} improve on NeRF by adding a local reflection model in addition to density. It yields impressive relighting results, albeit from single point light sources.

\item 
\href{https://nerf-w.github.io/}{NeRF-W}~\citep{MartinBrualla20arxiv_nerfw} is one of the first follow-up works on NeRF, and optimizes a latent appearance code to enable learning a neural scene representation from less controlled multi-view collections.
\end{itemize}

\section{Shape}

Latent codes can also be used to encode shape priors.

\begin{itemize}
\item 
\href{https://github.com/sxyu/pixel-nerf}{pixelNeRF}~\citep{Yu20arxiv_pixelNeRF} is closer to image-based rendering, where N images are used at test time. It is based on PIFu, creating pixel-aligned features that are then interpolated when evaluating a NeRF-style renderer.

\item
\href{https://github.com/alextrevithick/GRF}{GRF}~\cite{Trevithick20arxiv_GRF} is pretty close to pixelNeRF in setup but operates in a canonical space rather than in view space.

\item 
\href{https://autonomousvision.github.io/graf/}{GRAF}~\citep{Schwarz20neurips_graf} i.e., a “Generative model for Radiance Fields" is a conditional variant of NeRF, adding both appearance and shape latent codes, while viewpoint invariance is obtained through GAN-style training.

\item 
\href{https://marcoamonteiro.github.io/pi-GAN-website/}{pi-GAN}~\citep{Chan20arxiv_piGAN} is similar to GRAF but uses a SIREN-style implementation of NeRF, where each layer is modulated by the output of a different MLP that takes in a latent code.
\end{itemize}

\section{Composition}

It could be argued that none of this will scale to large scenes composed of many objects, so an exciting new area of interest is how to compose objects into volume-rendered scenes.

\begin{itemize}
\item 
\href{https://shellguo.com/osf/}{Object-Centric Neural Scene Rendering}~\citep{Guo20arxiv_OSF} learns "Object Scattering Functions" in object-centric coordinate frames, allowing for composing scenes and realistically lighting them, using Monte Carlo rendering.

\item
\href{https://arxiv.org/abs/2011.12100}{GIRAFFE}~\citep{Niemeyer20arxiv_GIRAFFE} support composition by having object-centric NeRF models output feature vectors rather than color, then compose via averaging, and render at low resolution to 2D feature maps that are then upsampled in 2D.

\item
\href{https://arxiv.org/abs/2011.10379}{Neural Scene Graphs}~\citep{Ost20arxiv_NSG} supports several object-centric NeRF models in a scene graph.
\end{itemize}

\section{Pose Estimation}

Finally, at least one paper has used NeRF rendering in the context of (known) object pose estimation.

\begin{itemize}
\item
\href{http://yenchenlin.me/inerf/}{iNeRF}~\citep{YenChen20arxiv_iNeRF} uses a NeRF MLP in a pose estimation framework and is even able to improve view synthesis on standard datasets by fine-tuning the poses. However, it does not yet handle illumination.
\end{itemize}

\section{Concluding Thoughts}

Neural Volume Rendering and NeRF-style papers have exploded on the scene in 2020, and the last word has not been said. This note definitely does not rise to the level of a thorough review, but we hope that an annotated bibliography is useful for people working in this area or thinking of joining the fray.

However, it is far from clear -even in the face of all this excitement- that neural volume rendering is going to carry the day in the end. While the real world does have haze, smoke, transparencies, etc., in the end, most of the light is scattered into our eyes from surfaces. NeRF-style networks might be easily trainable because of their volume-based approach, but we already see a trend where authors are trying to discover or guess the surfaces after convergence. In fact, the stratified sampling scheme in the original NeRF paper is exactly that. Hence, as we learn from the NeRF explosion we can easily see the field moving back to SDF-style implicit representations or even voxels, at least at inference time.

\bibliographystyle{apalike}
\bibliography{NeRF-and-Beyond}

\end{document}